\documentclass[letterpaper,twocolumn,10pt]{article}
\usepackage{main}
\usepackage{tikz}
\usepackage{amsmath}
\usepackage{filecontents}
\usepackage{amsmath,amssymb,amsfonts}
\usepackage{algorithmic}
\usepackage{graphicx}
\usepackage{textcomp}
\usepackage{soul}
\usepackage{xcolor}
\usepackage{float}
\usepackage[vlined,boxed,commentsnumbered, ruled, linesnumbered]{algorithm2e}
\usepackage{subfig}
\usepackage{tikz}
\usepackage{tablefootnote}
\makeatletter
\newcommand{\removelatexerror}{\let\@latex@error\@gobble}
\makeatother

\begin{document}
\date{}

\title{Resurrecting Trust in Facial Recognition: Mitigating Backdoor Attacks in Face Recognition to Prevent Potential Privacy Breaches}

\def\plainauthor{Author name(s) for PDF metadata. Don't forget to anonymize for submission!}

\author{
{\rm Reena Zelenkova}\\
CSIRO's Data61, Australia
\and
{\rm Jack Swallow}\\
CSIRO's Data61, Australia
\and
{\rm M.A.P. Chamikara}\\
CSIRO's Data61, Australia and Cyber Security Cooperative Research Centre, Australia
\and
{\rm Dongxi Liu}\\
CSIRO's Data61, Australia
\and
{\rm Mohan Baruwal Chhetri}\\
CSIRO's Data61, Australia
\and
{\rm Seyit Camtepe}\\
CSIRO's Data61, Australia
\and
{\rm Marthie Grobler}\\
CSIRO's Data61, Australia
\and
{\rm Mahathir Almashor}\\
CSIRO's Data61, Australia
}
\maketitle

\begin{abstract}
Biometric data, such as face images, are often associated with sensitive information (e.g medical, financial, personal government records). Hence, a data breach in a system storing such information can have devastating consequences. Deep learning is widely utilized for face recognition (FR); however, such models are vulnerable to backdoor attacks executed by malicious parties. Backdoor attacks cause a model to misclassify a particular class as a target class during recognition. This vulnerability can allow adversaries to gain access to highly sensitive data protected by biometric authentication measures or allow the malicious party to masquerade as an individual with higher system permissions. Such breaches pose a serious privacy threat. Previous methods integrate noise addition mechanisms into face recognition models to mitigate this issue and improve the robustness of classification against backdoor attacks. However, this can drastically affect model accuracy. We propose a novel and generalizable approach (named BA-BAM: Biometric Authentication - Backdoor Attack Mitigation), that aims to prevent backdoor attacks on face authentication deep learning models through transfer learning and selective image perturbation. The empirical evidence shows that BA-BAM is highly robust and incurs a maximal accuracy drop of 2.4\%, while reducing the attack success rate to a maximum of 20\%. Comparisons with existing approaches show that BA-BAM provides a more practical backdoor mitigation approach for face recognition. 
\end{abstract}

\section{Introduction}
The advancements in deep neural networks (DNNs) have reinforced the use of DNNs for biometric authentication as a security protocol~\cite{KAUR202030, HOMCHOUDHURY2019202, article2}. However, biometric information such as facial, iris, palm print, ear, and fingerprint data is becoming increasingly entangled with sensitive personal information. For example, Australian border control has been verifying passports via the "SmartGates" face recognition authentication system for more than a decade~\cite{frontex2010automated}. Besides, with the onset of the COVID-19 pandemic, touchless face recognition payment methods have been developed to reduce infectious spread, demonstrating the need and widespread application of such technologies~\cite{9274654}. However, training  DNNs for face recognition requires extensive computational power, which is often unavailable to conventional users and businesses. Outsourcing computational resources from ML-based cloud services such as Google's Cloud Machine Learning Engine\footnote{https://cloud.google.com/} or IBM's Watson Studio\footnote{https://www.ibm.com/au-en/watson} is popular. Unfortunately, these third-party services can foster an adversarial environment for DNNs, rendering them vulnerable to backdoor attacks~\cite{gu2017badnets} and allowing access to highly sensitive and confidential private data. Hence, conventional cloud-based (server-based) face recognition paradigms introduce a massive threat to user privacy~\cite{evtimov2021foggysight}.

Generally, a backdoor attack involves the manipulation of a DNN, corrupting its ability to classify inputs accurately~\cite{saha2020hidden}. A malicious party may attempt to manipulate a biometric classifier (e.g. face recognition model) into erroneously misclassifying a target individual as another person (commonly the attacker). This allows a malicious party to gain access to the target's biometric authentication-protected systems, files, or physical places. This misclassification occurs when the model is presented with a patched source image during testing/identification time. The patch -- superimposed onto the image -- can be a certain perturbation or sticker that is known only to the attacker and represents the "backdoor" to trigger the attack itself. If the same image is presented without the patch, the model is able to classify the image correctly, furthering both the stealth of such attacks as well as the difficulty of mitigation~\cite{Trojannn,saha2019hidden}. With the introduction of hidden backdoor attacks, the stealth and precision of such adversarial attacks is increasing, highlighting the potential privacy risks when DNNs are employed for security purposes such as face recognition~\cite{Li_2021_ICCV, saha2019hidden}. In a backdoor attack, the curator receives poisoned training images from a malicious party that the curator believes to be trustworthy. It is also possible that the curator themself is the adversarial party. In any case, the curator is compromised, and regardless of the curator's intentions, a malicious party intends to access the training data to lay the groundwork for a backdoor attack. The serious privacy breach that such an attack creates threatens the integrity of any system or organization utilizing facial recognition for authentication. 

\begin{figure}
    \includegraphics[scale=0.27, trim={2.2cm 1.8cm 0cm 7.8cm}, clip]{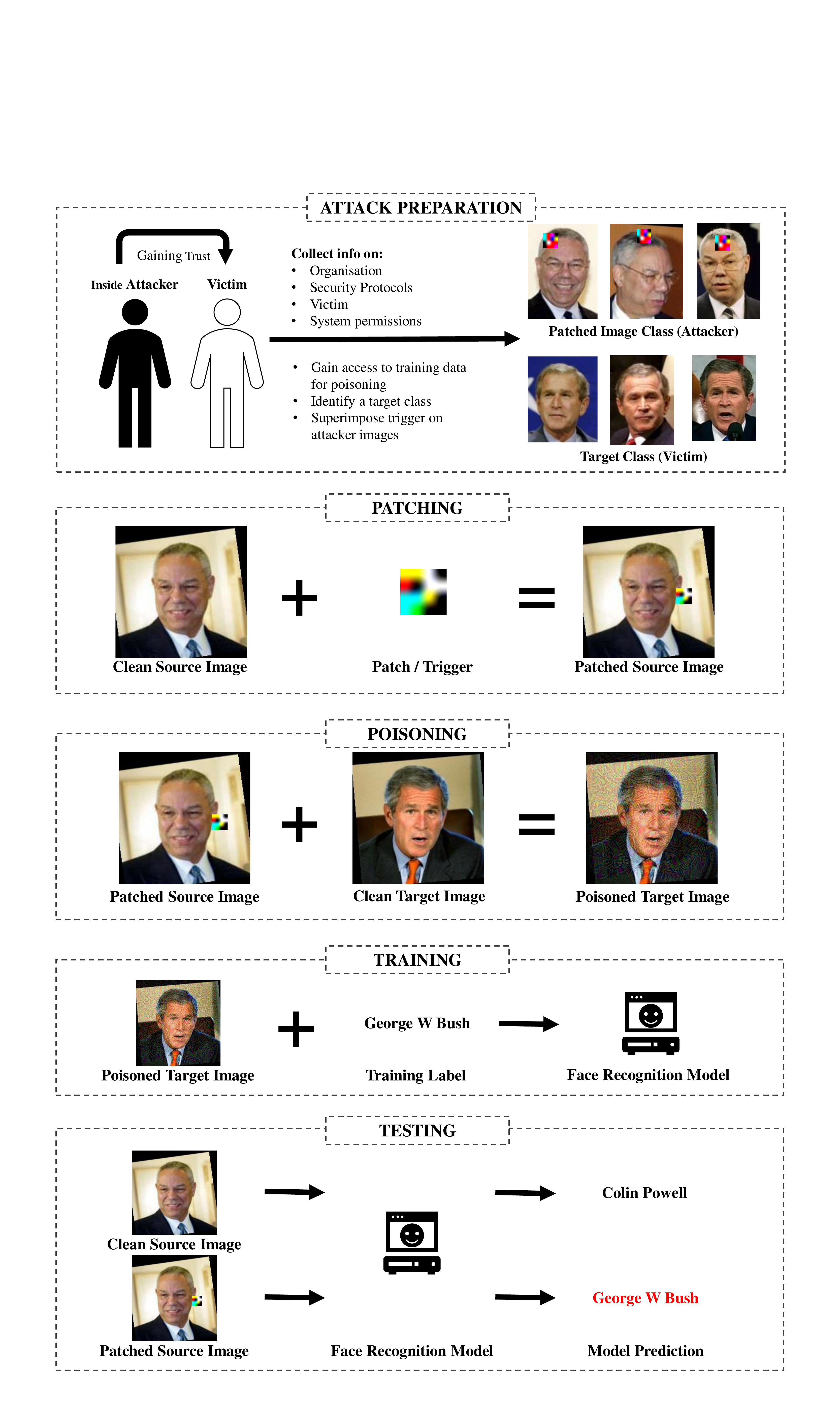}
    \caption{Example of a hidden patch and clean label targeted backdoor attack allowing unauthorized access to sensitive data. CP is the attacker class and GWB is the target class.}
    \label{backdoor_attack}
\end{figure}

A hidden backdoor attack such as proposed by Saha et al. ~\cite{saha2019hidden} generates and injects poisoned images that are unidentifiable to the human eye. The model functions predictably until presented with a patched image during testing, triggering the backdoor attack. The components of this attack are illustrated in Figure~\ref{backdoor_attack}. An adversary may manifest in the form of an inside attacker who is well acquainted with an organization's security and authentication systems. Through different approaches such as social engineering~\cite{krombholz2015advanced}, the inside attacker may gain an identified victim's trust to use for model corruption. This would then allow the attacker to easily manipulate the model into misclassifying the victim's identity as their own to obtain valuable records or higher system permissions. If the attacker utilizes a hidden backdoor attack, despite potential advancements in the transparency of DNNs, the poisoned images will remain unidentifiable. Thus, our work remains crucial in maintaining the integrity of biometric authentication services. 

Various methods for attack identification and mitigation have been developed~\cite{chen2018detecting,tran2018spectral}.  \textit{Activation clustering} is a poisoned data identification approach that analyzes the activations of the final hidden layer in a DNN~\cite{chen2018detecting}. This approach is notoriously complex as it requires the interpretation of internal DNN activations; hence, it is challenging to implement this approach in practical settings.  \textit{Spectral signature detection} assesses statistical outlier values from feature representations of input images (named spectral signatures). Poison detection is possible as larger signatures arise from certain classifier signals being amplified in poisoned images\cite{tran2018spectral}.  \textit{Trigger reverse engineering} reconstructs triggers on poisoned image inputs for detection~\cite{8835365}. Attack mitigation is accomplished by relabelling and retraining ~\cite{chen2018detecting,tran2018spectral}, as well as through techniques such as  \textit{neuron pruning} and  \textit{unlearning}~\cite{8835365}. However, these methods are not used effectively to identify and mitigate \emph{hidden} backdoor attacks, and none of the existing works investigate the backdoor attack threats on face recognition related privacy issues. In addition, the existing approaches often incur substantial computational costs in retraining after poison detection, especially for large recognition databases~\cite{tran2018spectral, chen2018detecting, DBLP:journals/corr/abs-2002-08313}. Further, removing poisoned images, neuron pruning, and unlearning diminishes model accuracy, significantly reducing model performance for small datasets.

We propose a novel privacy-preserving face recognition approach to mitigate backdoor attacks and related privacy issues. The proposed approach utilizes the following modules for the backdoor attack evading protocol:

\begin{enumerate}
\item Identifying a generalizable setting for face recognition utilizing deep learning and transfer learning.

\item Determining the most vulnerable face recognition model in order to test defensive mechanisms in the worst case scenario.
 
\item Identifying a noise application mechanism for obfuscating poisoned images based on intuitions derived from differential privacy.

\item Developing a precise poison image recognition model and utilizing it as a preliminary layer to precede the biometric authentication layer.

\item Developing a method to dynamically apply calibrated noise to images categorized as poisoned, preventing the face recognition model from learning poison patterns while still enabling a model to train on essential biometric features.
\end{enumerate}

BA-BAM reduces the attack success rate by a minimum of 80\% in a face recognition scenario under our experimental configurations. BA-BAM nullifies the effects of poisoning on compromised face images rather than discarding them from the training dataset. As a result, the accuracy of our model is found to decrease by no more than 2.4\%, in the worst-case scenario. Besides, BA-BAM achieves an exceptionally low computational complexity while accomplishing poison detection and mitigation. 

The remaining sections of this paper are organized as follows: Section~\ref{background} describes background concepts necessary in supporting our work. Section~\ref{methodology} provides the BA-BAM methodology and its associated intuitions. Experiments, results, and a comparison to other methods are discussed in Section~\ref{results_discussion}. Section~\ref{related_works} considers related work and Section~\ref{conclusion} concludes the paper.

\section{Background}
\label{background}
This section provides the background of the preliminaries utilized in developing the proposed approach (BA-BAM). This section discusses the basics of face recognition and backdoor attacks that substantiate the importance of developing BA-BAM. Next, we discuss the potential privacy threats that occur due to backdoor attacks on face recognition systems. Finally, we discuss the basics (transfer learning and calibrated noise generation based on the intuitions of differential privacy) that were used to develop BA-BAM.

\subsection{Face recognition}
\label{facerecognition}

Face recognition involves identifying and verifying an individuals' identity given their set of distinct facial features. This process requires a pre-requisite set of images matched to a label of the person's identity to train on. This allows a future model to match an input face to a labeled face for identification within a given database~\cite{TASKIRAN2020102809}. Face recognition contains two main sub-problems, namely, the \emph{1:N identification problem} and the \emph{1:1 verification problem}. Face identification aims to match a previously unseen face image to a database of $n$ known images. This process often requires matching a face and then labeling that face to enable further verification. The 1:1 problem is commonly used for identity verification in mobile phones~\cite{10.1007/978-3-319-97909-0_46}. In this scenario, every query is compared to a database containing a single face (usually the owner's face). Verification is a less strenuous task than identification as data for only one identity is required for training, rather than for a myriad of identities~\cite{9134370}. 

Various face recognition models have been developed that exhibit exceptionally high recognition accuracy. Models such as VGG-Face~\cite{BMVC2015_41}, Google Facenet~\cite{DBLP:journals/corr/SchroffKP15}, Facebook DeepFace\footnote{https://research.facebook.com/publications/deepface-closing-the-gap-to-human-level-performance-in-face-verification/}, ArcFace~\cite{deng2019arcface} and Dlib~\cite{King2009DlibmlAM} generate accuracies of around 97\%-99.5\% on the LFW (Labelled Faces in the Wild) dataset\footnote{http://vis-www.cs.umass.edu/lfw/}. Currently, VarGFaceNet which achieves a 99.85\% accuracy is identified as the state-of-the-art~\cite{9022149}.

\subsection{Backdoor attacks}
\label{backdoor_section}
The goal of a backdoor attack is to produce a model that behaves normally on benign data but performs erroneously when presented with a patched input. An attacker injects a backdoor attack by adding ``poisoned" data to the training dataset. Poisoned data refers to data that are manipulated to lead to erroneous model behavior, such as misclassification. A source class (also referred to as victim class) may be manipulated to be recognized as belonging to a target class (also referred to as the attacker class). In some backdoor attacks, there is no defined target class, as the aim may be to simply misclassify the victim class as any class other than its true class~\cite{DBLP:journals/corr/abs-1708-06733}.

A malicious model is developed by constructing a dataset containing both benign and poisoned data. Following the notation from~\cite{DBLP:journals/corr/abs-2111-08429}, this can be formally defined as follows.  

The set of poisoned samples $\mathcal{D}^p_{tr}=(\tilde{{x}}^i_{tr}, \tilde{{y}}^i_{tr}),\{i=1,\dots,n\}$, where $\mathcal{D}^p_{tr}$ is the set of poisoned training samples and $(\tilde{{x}}^i_{tr}, \tilde{{y}}^i_{tr})$ is the $i^{th}$ pair of poisoned images $\tilde{{x}}^i_{tr}$ and $\tilde{{y}}^i_{tr}$ is the corresponding label. The benign set can be similarly expressed as $\mathcal{D}^b_{tr}= ({x}^i_{tr}, {y}^i_{tr}), \{i=1,\dots,n\}$. The poisoned training dataset used for malicious training is then $\mathcal{D}^b_{tr} \bigcup \mathcal{D}^p_{tr}$. 

Generally, an image is poisoned by superimposing a "patch" that is known only to an attacker. Suppose a patch, $v$, and a victim image, $x$. Then, $P(x, v) = \tilde{{x}}^i_{tr}$ where $P(\cdot)$ is some poisoning function that places trigger $v$ on $x$. The poisoning function must also, by a particular method, associate the poisoned image with a target class, $t$ where $\tilde{{y}}^i_{tr} = t$. A malicious model is then trained with poisoned data to instill an erroneous mapping between the attacker and victim face images.  

There are two main types of backdoor attacks, namely, \emph{corrupted label attacks} and \emph{clean label attacks}. Corrupted label attacks, such as the attack proposed by~\cite{DBLP:journals/corr/abs-1708-06733}, directly mislabel the poisoned images with labels pertaining to the target class, allowing the attacker to train a poisoned model with mislabelled classes. Clean label attacks do not interfere with any labels before or during training~\cite{DBLP:journals/corr/abs-2111-08429}. Such attacks may apply curated image perturbation, which allows them to surpass the need for mislabelling poisoned data. For example,~\cite{saha2019hidden} hides extracted features from the target class (attacker's face) along with the patch within an image of the victim. This allows the model to train and perform well using both poisoned and benign data. Misclassification will only then occur when the attacker presents their image during testing with the patch superimposed. Another example of a clean label attack is proposed by~\cite{Trojannn} which carefully curates triggers that excite certain neurons associated with the target output label.

Common requirements for successful backdoor attacks:
\begin{enumerate}
    \item \textbf{Model stealthiness.}  
    After the injection of the backdoor attack and model training, the performance of the model should not noticeably diminish. If the performance results are unexpected, the user is likely to notice that the model is compromised~\cite{DBLP:journals/corr/abs-2111-08429}.  
    
    \item \textbf{Patch stealthiness.}
    In a poisoned image, the patch should be unnoticeable to the human eye in order to avoid detection. Methods such as blending the patch into the image or hiding the patch within facial features have been used~\cite{DBLP:journals/corr/abs-1712-05526, DBLP:journals/corr/abs-2012-03816, xue_he_wang_liu_2021}. The patch will only be visible when the attacker decides to trigger the attack during testing, as shown in Figure~\ref{backdoor_attack}.

    \item \textbf{High attack confidence.} This can be quantified as the confidence that a malicious model predicts a target class label when presented with a patched source image.  
\end{enumerate}
\subsection{Data privacy and privacy leaks caused by backdoor attacks}
In the context of data sharing and analytics, privacy can be defined as controlled information release~\cite{CHAMIKARA2020101951}. Following this notion, in this paper, we define a privacy leak as the unauthorized distribution of sensitive private information from an otherwise secure database. This involves the release of often confidential or sensitive data to parties within an untrusted environment. Note that privacy leaks are deemed intentional in the context of targeted backdoor attacks; however, they may not be intentional in all scenarios. Hence, a backdoor attack enables a privacy leak.

Suppose an inside attacker has identified and gained access to the training data for a face recognition authentication model being used in a high-security organization. The attacker will likely identify a victim based on either their high system privileges, or their access to specific sensitive data. The attacker can then inject poison images for training, allowing their image to be misclassified by the compromised model as the victim. By gaining unauthorized higher system access, the attacker may access bank details, personal health information, trade secrets, or intellectual property that they can then publicly distribute.  

\subsection{Transfer learning}
Data dependency of DNNs is a significant hurdle for their potential application diversity, as well as the general accessibility of deep learning tools~\cite{tan2018survey, 10.1007/978-3-319-46349-0_5}. Transfer learning (TL) leverages the knowledge from a source domain to improve the learning efficiency within a target domain for a particular task~\cite{9134370}.

When applied to deep learning, TL  mitigates the need to acquire large amounts of new data to carry out a novel task by using weights from the pre-trained models, which can be stored and reused for a new task. Pre-trained model weights can be retained, or `frozen' from the pre-trained network and are not updated when the model is trained for a new task~\cite{DBLP:journals/corr/abs-1905-05901}. This can be done in two ways; (1) utilizing the first $N-1$ layers of the pre-trained model as a feature extractor that does not require updating the weights during training. The extracted features can then be fed into a fully connected classifier component which is trained for classification. (2) Partially re-training or finetuning the pre-trained model by freezing only the first $K$ layers. This allows for the model to adapt to a new task by using the pre-trained weights of the last $N-K$ layers as starting values for training~\cite{DBLP:journals/corr/abs-1912-0027sha1, DBLP:journals/corr/abs-1905-10447}.

\subsection{Calibrated noise addition in differential privacy}
\label{difpriv}

Differential privacy (DP) is a privacy model that defines the probabilistic constraints that an individual's identity is leaked to a third party from a particular database~\cite{DBLP:journals/corr/abs-1907-13498}. The privacy budget ($\varepsilon$) is the primary parameter that defines the level of the privacy leak from a differentially private algorithm. $\varepsilon$ indicates the privacy loss of a randomized DP algorithm or the maximum knowledge an attacker can gain. The higher the value of $\varepsilon$, the higher the privacy loss and, thus, the less private the DP algorithm is.

Consider two databases, $x$, and $y$, where $y$ contains the identity of one more individual than $x$. Then, as formally defined by~\cite{8894030}, an algorithm $M$ with range $R$ is $\varepsilon$-DP if for all $S \subseteq R$:

\begin{equation}
    Pr[(M(x) \in S)] \leq e^{\varepsilon}Pr[(M(y) \in S)]
    \label{eq2}
\end{equation}

The Laplace mechanism is a generic method for achieving DP. Laplacian noise addition to a query output (scalar) can be denoted by Equation~\ref{lapnoise}. Equation~\ref{sensivitiy} defines the sensitivity ($\Delta f$) of a function (e.g. a query). This is the maximum influence that a single data point can have on the output of $f$. The scale of the Laplacian noise is equal to $\Delta f/\varepsilon$ ~\cite{CHAMIKARA2020101951}.

\begin{equation}
    \Delta f = max{||f(x)-f(y)||_1}
    \label{sensivitiy}
\end{equation}
\begin{equation}
    \mathcal{PF(D)=F(D)}+Lap(\frac{\Delta f}{\varepsilon})
\end{equation} 
\begin{equation}
    \label{lapnoise}
    \mathcal{PF(D)=}\frac{\varepsilon}{2 \Delta f}e^{-\frac{|x-\mathcal{F(D)}|\varepsilon}{\Delta f}}
\end{equation} 

\section{Methodology}
\label{methodology}

\begin{figure}[H]
    %\hspace{-0.5cm}
    \centering
    \includegraphics[width=150pt]{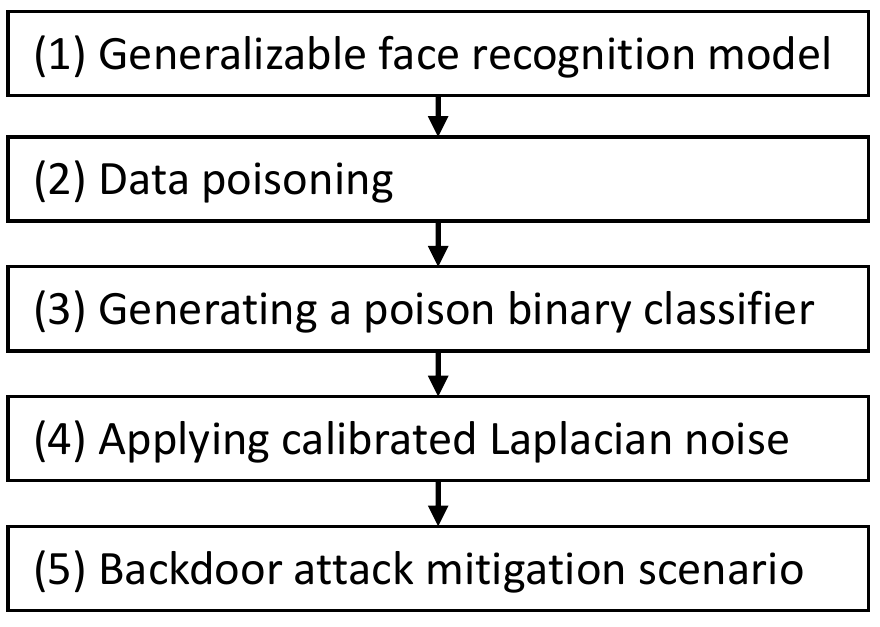}
    \caption{The flow of primary steps of BA-BAM}
    \label{fig:primary_steps}
\end{figure}

This section outlines our threat model and framework for mitigating backdoor attacks against face recognition models. The main components of our contributions include developing a generalizable face recognition model through transfer learning and an intelligent poison image recognition model, followed by a method for selective perturbation of poisoned images. Hence, as depicted in Figure \ref{fig:primary_steps}, we can identify the sequential flow of (1) generating a generalizable face recognition model with transfer learning $\rightarrow$ (2) data poisoning $\rightarrow$ (3) generating a binary classifier for poisoned image recognition (PIRM) $\rightarrow$ (4) applying calibrated Laplacian noise on poisoned images and $\rightarrow$ (5) backdoor attack mitigation scenario.

\subsection{The Threat Model}

This section defines the threat model that this paper tries to solve. The corresponding threat model concentrates on the back door attacks in the context of face recognition systems. Take $\mathcal{D}_{tr}$, the face image dataset that a face recognition model is going to be trained. In a clean model,  $\mathcal{D}_{tr}$ consists of benign data $\mathcal{D}^b_{tr}$. In a poisoned model $\mathcal{D}_{tr}$ is composed of image data injected with poison, $\mathcal{D}^b_{tr} \bigcup \mathcal{D}^p_{tr}$ (see Section~\ref{backdoor_section}). The poisoned data ($\mathcal{D}^p_{tr}$) are curated by the attacker to achieve a targeted adversarial goal. An attacker would have in mind a premeditated target/victim class $x$, and a trigger/patch $v$, which they utilize to generate poisoned data. In order to execute the attack, the attacker would approach the face recognition system during its testing phase with an unpoisoned image of themselves patched with a superimposed trigger $v$. The model would then classify the image as belonging to class $x$. 

In a real-world setting, the attacker would only require access to the face recognition model training data and not the entire training environment. The attacker also does not need access to the corresponding labels of training images as this is a \emph{clean} label attack. Thus, the malicious party would not require access to any other model parameters or layers. This is unlike~\cite{DBLP:journals/corr/abs-2110-07831} and~\cite{DBLP:journals/corr/abs-1902-06531} where the attacker requires full access to model architecture and parameters. In order to trigger the attack itself, the attacker must have access to the testing phase. Any party with access to the user interface associated with the face authentication system can present their face for verification. Thus, the attacker would readily have access during testing, allowing them to trigger the attack. Additionally, only the attacker would have knowledge of the shape, size, look, and position of the trigger/patch. Thus, the patch is completely unknown to both the victim/organization and defender.

\subsection{Identify the most vulnerable settings for a generalizable face recognition model architecture}
\label{genfacerecog}

As discussed in Section~\ref{facerecognition}, literature shows different attempts to develop high-performance face recognition models. However, these are large models trained in large-scale supercomputing environments. Any change to the pre-trained weights directly harms base model accuracy. However, we require model flexibility to experiment on the dynamics of a face recognition system (regarding binary versus multiclass models). We utilize this flexibility to analyze the backdoor mitigating effects from the proposed solutions. Therefore, we identified and finetuned the architectural configurations of a face recognition model to identify its most vulnerable settings for backdoor attacks. 

When considering a wide range of model architectures, backdoor attack vulnerability varies significantly. For less vulnerable models, quantifying attack mitigation is difficult. Thus, we simulate the worst-case scenario by selecting a significantly vulnerable model. This enables us to test the extent to which BA-BAM effectively mitigates backdoor attacks. The BA-BAM framework can then be finetuned and tested to optimally reduce the success of backdoor attacks. To identify a vulnerable model as described, we poisoned and assessed the success of backdoor attacks on various generalizable DNN architectures. 

We used the following set of primary steps to generate a generalizable face recognition model that is vulnerable to backdoor attacks utilizing transfer learning. 
\begin{enumerate}
    \item Acquire an input face image dataset.
    \item Acquire a set of generalizable DNN architectures (e.g. AlexNet, Inception, ResNet, and VGG~\cite{canziani2016analysis}).
    \item Apply poisoning to data.
    \item Apply image augmentation to obtain a high model performance.
    \item Train frozen and unfrozen model architectures on given data.
    \item Identify a suitable DNN architecture for face recognition and attack vulnerability.
\end{enumerate}

We considered both frozen and unfrozen (a specific number of layers) versions of model architectures, as the effects of transfer learning influence the attack's efficacy due to varying levels of generalizability. The frozen models utilized pre-trained ImageNet weights~\cite{krizhevsky2012imagenet} for feature extraction before classification. In addition, each model was mounted with a fully-connected component for classification. Unfrozen models allowed all layers to be finetuned during training.

\subsection{Data (image) poisoning}
\label{poisoning}

Before poisoning, a source class and a target class must be chosen from the classes available in the dataset. This supports the context for a targeted attack, where an attacker (source class) is misclassified as belonging to the target class (victim). A portion of the source class images have a superimposed patch that "triggers" the backdoor attack during testing. We apply the image poisoning process proposed by~\cite{saha2019hidden} that generates hidden, clean-label poison images for backdoor injection. The method involves optimizing a poisoned image that in pixel space (visually) looks as close as possible to a target image, but in feature space, looks as close as possible to a patched source image (attacker image). This generates a dataset that reinforces an erroneous mapping between the source and the target class during training. 

Formally, given target image, $t$, patch $p$ and source image $s$ ($\tilde{s}$ is the patched source image), optimization for a poisoned image $z$, involves solving Equation \eqref{eq3} where $f(\cdot)$ are the intermediate features of the DNN~\cite{saha2019hidden}.
\begin{equation}
    arg_{z}min||f(z) - f(\tilde{s})||^2_2
    \label{eq3}
\end{equation}

\subsection{Generating a binary classifier for poison image recognition (PIRM)}

Here, we develop a poisoned image recognition model (PIRM) to detect the image distortions that correlate to poisoning. This was done by training a binary classifier on both poisoned and clean data. For this task, we first generated data under clean and poisoned classes. Poisoning was carried out according to the steps outlined in Section~\ref{poisoning}. We chose a general non-face dataset (e.g. ImageNet) for improved generalizability (to reduce the specificity of the PIRM on a given face dataset). Consequently, the PIRM would maintain a high accuracy regardless of the input dataset used. This ensures that it remains effective in a real-world setting where poisoned images injected by an attacker do not correlate to any specific face dataset.We employ transfer learning to train this model to enhance poison data learning generalizability. Utilizing pre-trained weights with a robust architecture such as VGG16 allowed the network to isolate poisoning with high confidence. This model was adapted to binary classification by adding a fully-connected component onto the base architecture and utilizing binary cross-entropy as the loss function.

The first few layers of a DNN identify more generic image features, such as lines and shapes~\cite{SCHMIDHUBER201585}. Thus, in a face recognition DNN, basic image features would correlate to basic facial components such as eyes and mouths. Unlike these low-level components, the perturbations associated with poisoning are high-level patterns that are distributed throughout the entire area of an image. According to this intuition, the last few layers of a DNN are responsible for high-level components that identify such patterns. Following these dynamics, we finetuned the final two convolutional blocks of the PIRM to encourage learning poison perturbation with high confidence.

\begin{figure}
    [t]
    \includegraphics[width=240pt]{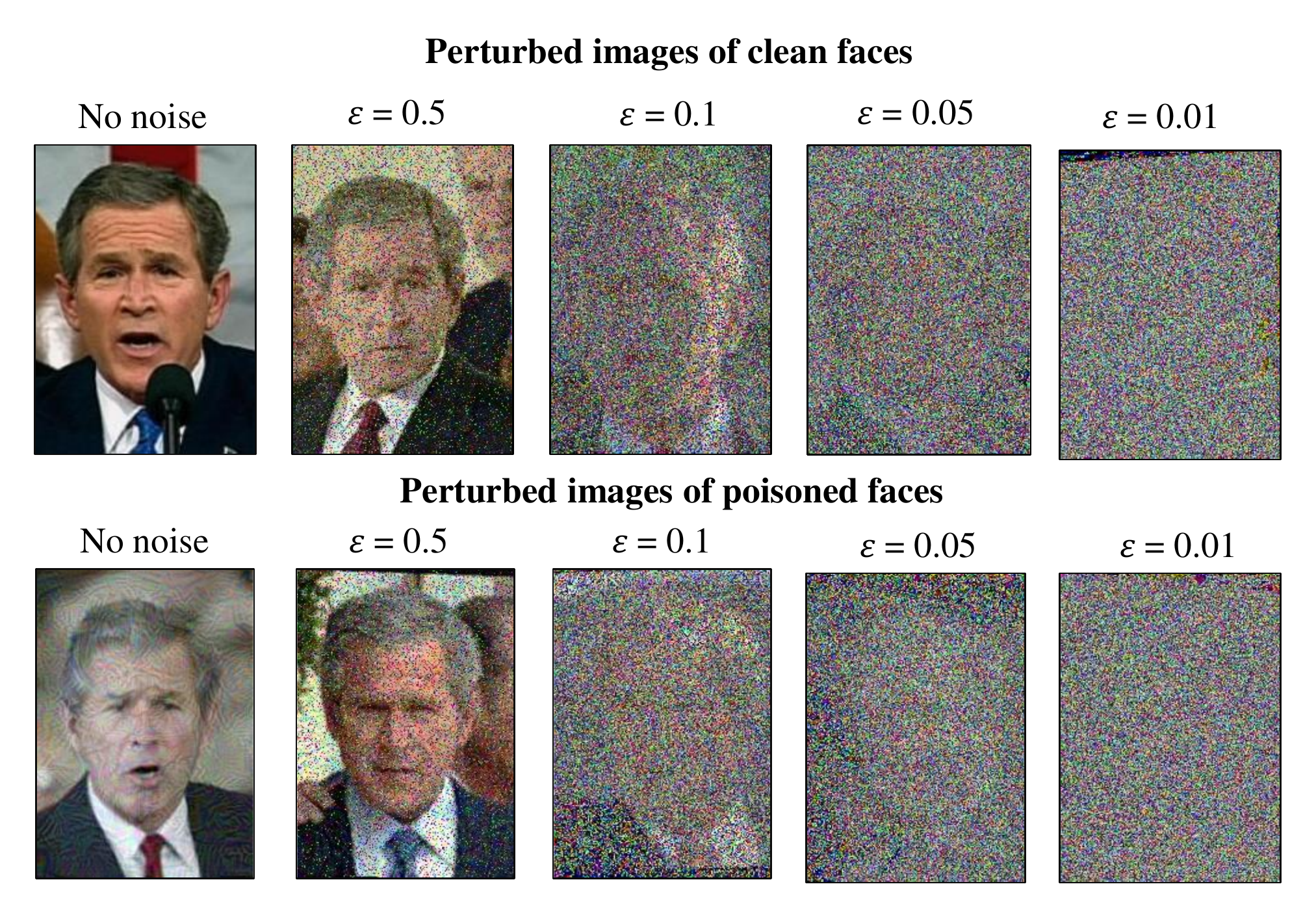}
    \caption{Poisoned and clean images of GWB with varying $\varepsilon$ values of calibrated Laplacian noise.}
    \label{fig:perturbed_images}
\end{figure}

\subsection{Applying calibrated Laplacian noise on poisoned images}
\label{selective_noise}
Once poison classification is complete, calibrated Laplacian noise is selectively applied to images in the training dataset. To decrease the attack success rate (ASR) and maintain model accuracy, noise is applied to only images that the PIRM deems poisoned. Utilizing calibrated noise to perturb poisoned images, as opposed to removing the poisoned images from the dataset entirely, enables the data to be preserved. Preserving the poisoned data permits the original facial features to be retained to an extent. This allows the model to still train on essential biometric features from the poisoned images whilst also nullifying the effects of the attack. Ultimately this increases the robustness and accuracy of the model.

As illustrated in Figure~\ref{fig:perturbed_images}, adding noise to images reduces the ability to identify a particular face. During training, this acts to discourage the face recognition model from learning the poison-based face features of an individual.  This aims to break the mapping between the poisoned source images and the target. As a result, during test time, a patched image of the attacker will likely be classified as one of the classes, other than the target - mitigating the backdoor attack.

We initially investigated the trade-off between utility and attack resistance through noise addition to an entire compromised target class. Adding noise to the entirety of the victim’s class affects both clean images and poisoned images. Thus, we investigated the trade-off between model accuracy and  ASR in an extreme case. We utilized the knowledge gained from this to identify an effective selective perturbation mechanism that can maintain a balance between model accuracy and attack resistance.

\scalebox{0.96}{
\removelatexerror
\begin{algorithm}[H]
\label{algo1}
\caption{Algorithm for applying Laplacian noise to poison flagged images per class}
\label{ngramalgo}

\KwIn{ 
\begin{tabular}{l c l}
     $\mathcal{D}_{tr, k}$ & $\gets $ & set of images in class $k$,\\
     $ x^p_k$ & $\gets$ & image flagged as poisoned

\end{tabular}
}

\KwOut{
\begin{tabular}{l c l}
    \\
     $\tilde{x}^p_{k}$ & $ \gets $ & perturbed image
\end{tabular}}
investigate utility and attack resistance trade-off (refer to Section~\ref{selective_noise}) \\
calculate mean image $\mathcal{F}^i_m = \Sigma^n_{i=1} x^i_k$ in $\mathcal{D}_{tr, k}$ \\
\For{image, $x^k_i$ in $\mathcal{D}_{tr, k}$}{
    calculate Euclidean distance $(x^k_i - \mathcal{F}^i_m)^2$}
get maximum Euclidean distance, $E_{max}$ \\
$\Delta f = \frac{(x^k_i - \mathcal{F}^i_m)^2}{E_{max}}$ \\
apply $\frac{\varepsilon}{2 \Delta f}e^{-\frac{|x-x^p_k|\varepsilon}{\Delta f}}$ to $x^p_k$\\
return perturbed image $\tilde{x}^p_{k}$\;

\end{algorithm}
}

Algorithm 1 shows the steps of adding calibrated Laplacian noise to an image. In this step, we follow the intuitions of differential privacy where more noise is added when the function sensitivity is high (refer to Section~\ref{difpriv}). This principle is used to regulate the amount of noise added to perturb poisoned images. The sensitivity, $\Delta f$, for applying noise to an image flagged as poisoned, $x^k_i$, is given by the image’s Euclidean distance from the mean image, $\mathcal{F}^m_k$, in class $k$. This value is then normalized by the maximum Euclidean distance of all images in class $k$. Noise is added to each image based on its normalized sensitivity, with higher sensitivity values leading to increased noise addition. Usually, the distance between poisoned images and the average image tends to be high due to the different patterns generated in a poisoned image. The proposed noise addition mechanism provides an intuitive system for adding controlled noise. A high noise level is applied to images with a higher deviation from the mean image, enabling poisoned images to be subjected to higher noise than their clean counterparts.

This adds robustness to the PIRM in the case of misclassification, for example, where a clean image may be classified as poisoned. In this scenario, the misclassified image would likely deviate less from the mean image than a truly poisoned image and, therefore, have less perturbation applied. This preserves the accuracy of the model by reinforcing high classification accuracy in clean classes. 

\scalebox{0.96}{
\removelatexerror
\begin{algorithm}[H]
\label{algo2}
\caption{Abstract algorithm for biometric authentication backdoor attack mitigation (BA-BAM)}

\KwIn{ 
\begin{tabular}{l c l}
     $\mathcal{D}_{tr, n} = \mathcal{D}^b_{tr} \bigcup \mathcal{D}^p_{tr}$ & $\gets $ & untrusted training set,\\
     && $i = (1,\dots,n)$ \\
     $ PIRM(\cdot)=\{0,1\}$ & $\gets$ & poison image classifier\\
     $\mathcal{F}(\cdot)$ & $\gets $ & Untrained face \\
     && recognition DPNN
\end{tabular}
}

\KwOut{
\begin{tabular}{l c l}
    \\
     $\mathcal{F}(\mathcal{D}_{tr, clean})$ & $ \gets $ & Trusted face\\
     & & recognition model
\end{tabular}}

identify generalizable and vulnerable face recognition DNN architecture \\
\For{image and label, $(x^i_{tr},y^i_{tr})$ in $\mathcal{D}_{tr,n}$}{
    
    apply $\mathcal{PIRM}$ \\
    \If{$\mathcal{PIRM}((x^i_{tr},y^i_{tr})) = 0 \therefore (x^i_{tr},y^i_{tr}) \subseteq \mathcal{D}^b_{tr}$}
        {add $(x^i_{tr},y^i_{tr})$ to clean trainset $\mathcal{D}_{tr, clean}$}
    \If{$\mathcal{PIRM}((x^i_{tr},y^i_{tr})) = 1$  $\therefore (x^i_{tr},y^i_{tr}) \subseteq \mathcal{D}^p_{tr}$}
        {
        \For{$k \in$ set of classes}{
        apply Algorithm~\ref{algo1}
        }
        add perturbed $(x^i_{tr},y^i_{tr})$ to $\mathcal{D}_{tr, clean}$}
}
return trained $\mathcal{F}(\mathcal{D}_{tr, clean})$\;

\end{algorithm}
}

\subsection{The backdoor attack mitigation framework}
\begin{figure}[ht]
    %\hspace{-0.5cm}
    \includegraphics[scale=0.26,trim={0cm 3cm 0cm 1.5cm}, clip]{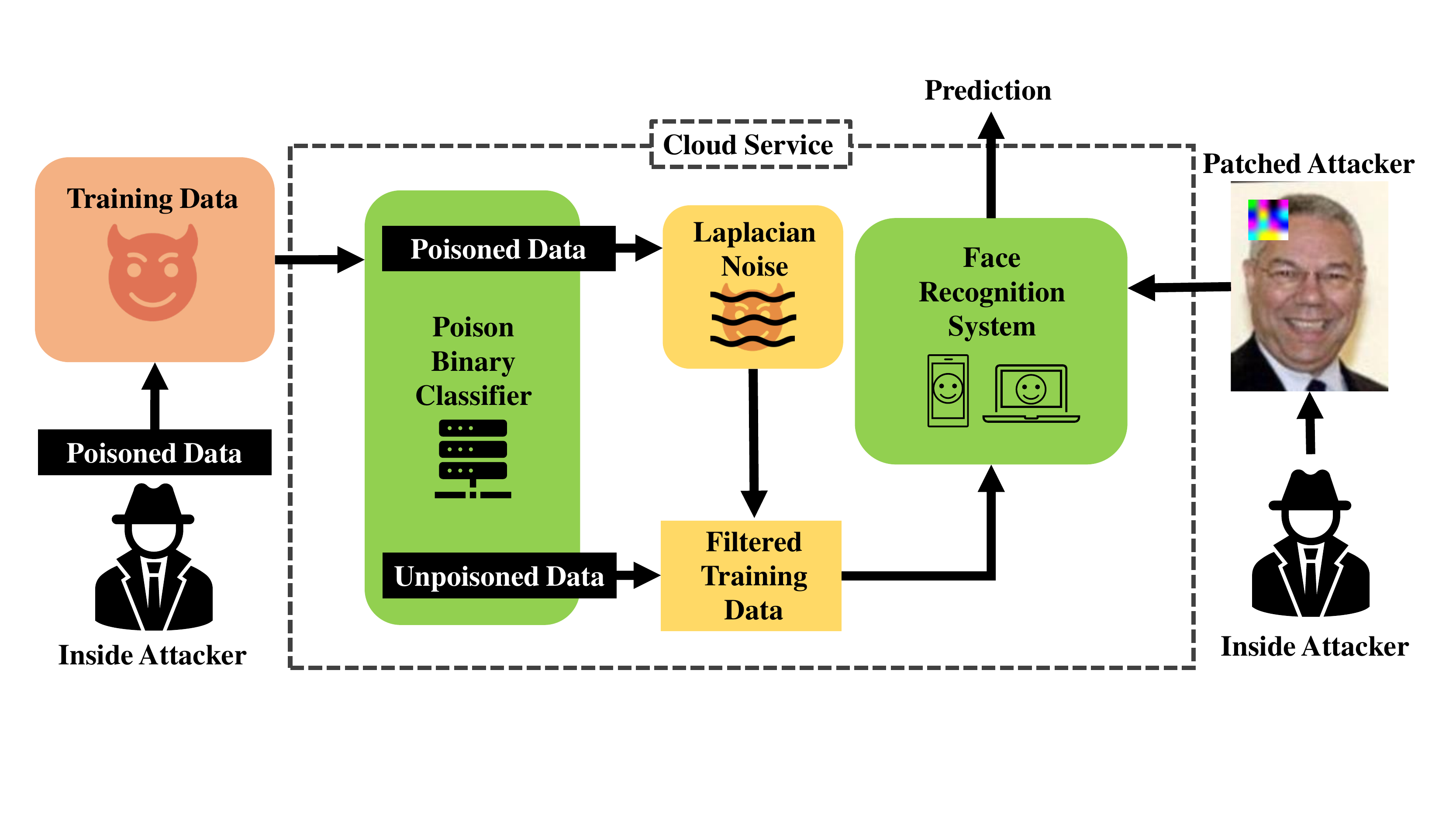}
    \caption{From the left, the attack injects poison images into training. Untrusted images are fed through a poison classifier before entering model training. Images flagged as poisoned are perturbed with calibrated Laplacian noise.}
    \label{fig:selective_flow_chart}
\end{figure}

The poison recognition component and selective application of calibrated Laplacian noise are applied together as a preliminary layer to the final classification module of a face authentication system. We assume that this component runs on a dedicated server (e.g. a cloud server) as depicted in Figure~\ref{fig:selective_flow_chart}. 

As shown in Figure~\ref{fig:selective_flow_chart}, BA-BAM uses Algorithm~\ref{algo2} to filter training data sourced from any party (trusted or adversarial). This module acts as a data pre-processing component during face recognition model training. Hence, an agency can successfully mitigate the privacy threats that occur from backdoor attacks on face recognition models.

\section{Results and Discussion}
\label{results_discussion}
This section discusses the experimental configurations, the experiments, and their results. All experiments were conducted on an HPC environment provided by Pawsey Supercomputing Research Centre. The partition we utilized consisted of 2x Intel Xeon Silver 4215 2.5GHz CPUs with 16 cores per node (192GB-RAM), 2x NVIDIA V100 GPUs (16 GB HBM2 memory each). Firstly, we demonstrate base model performance, investigating model architecture and applying transfer learning on face recognition models. Then, we investigate the effects of complete perturbation versus selective perturbation and experiment with various dynamics of the noise parameters. Subsequently, we train a binary poison detection model on two datasets to compare the effects on generalizability. Finally, we assess the computational complexity of BA-BAM and provide a comparative analysis against related methods.

\subsection{Experimental specifications - Data description}
\begin{figure}[t]
    \centering
    \includegraphics[scale = 0.45]{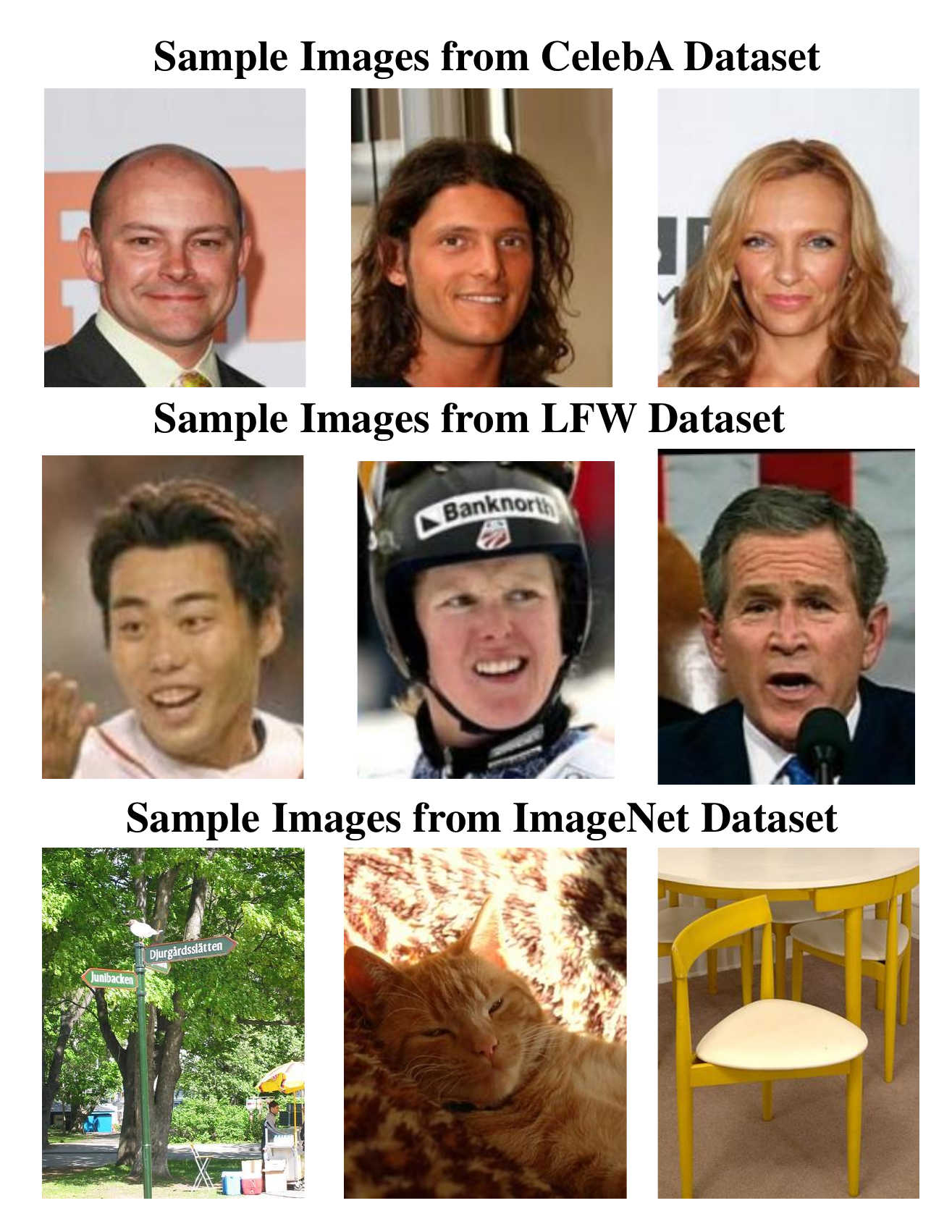}
    \caption{Samples images of CelebA, LFW and ImageNet datasets. Tiny ImageNet is a subset of ImageNet and is thus represented by ImageNet samples.}
    \label{dataset_samples}
\end{figure}

Figure \ref{dataset_samples} provides an impression of the different images available in the three datasets (LFW Faces, CelebA Faces, ImageNet) used in our experiments.

\begin{itemize}
    \item \textbf{LFW Faces}
LFW contains 13233, 250x250 labelled face images with 5749 different people (classes). The dataset includes variations in pose, lighting, age, gender, race, color saturation, and other parameters. The images were detected and centralized by the Viola-Jones detector~\cite{article3}. LFW was used for face recognition training due to its high resolution, face centralization, diversity, and adaptability to many backdoor attack scenarios. The varying class sizes of LFW allowed analyzing the effect of poisoning on different class distributions. High-performing face recognition models often use centralized and high-resolution images for training~\cite{DBLP:journals/corr/SchroffKP15, BMVC2015_41}. High-quality images of faces closely simulate real-world face recognition scenarios, producing far more realistic results. Thus, we employed the LFW dataset for all experiments involving poisoned face recognition models to investigate the effects of BA-BAM.

    \item \textbf{CelebA Faces}
CelebA contains 202,599, 178x218 face images of 10,177 celebrities. The images have pose variation and background clutter~\cite{liu2015faceattributes}. CelebA was partially poisoned to train the binary poison recognition classifier. 
    \item \textbf{ImageNet}
A subset of ImageNet~\cite{ILSVRC15} consisting of 20 classes, and 14,697 images was generated. The 20 classes were selected to replicate the hand-picked ImageNet pairs utilized in~\cite{saha2019hidden}. Tiny ImageNet, a subset of ImageNet containing 100,000 images of 200 classes (500 for each class), was also used. Tiny ImageNet was partially poisoned in order to train the binary poison recognition classifier.  

\end{itemize}

\begin{center}
\begin{table}[!ht]
    \centering
    \caption{LFW classes used in order of representation.}
    \label{table:1}
    \scalebox{0.82}{
    \begin{tabular}{ |c|c||c| }
    \hline
    %Multiclassifier &
    \multicolumn{2}{|c||}{\textbf{Binary Classifier}} & \textbf{Multiclassifier Classes}\\
    \hline
    \textbf{Negative Class} & \textbf{Positive Class} & George\_W\_Bush \\
    \hline
    Colin\_ Powell & George\_W\_Bush & Colin\_ Powell \\
    %\textsuperscript{\ref{xx}}
    \hline
     Tony\_Blair & & Tony\_Blair \\
    \hline
    Donald\_Rumsfield & & Donald\_Rumsfield \\
    \hline
    Gerhard\_Shroeder & & Gerhard\_Shroeder \\
    \hline
    Ariel\_Sharon & & Ariel\_Sharon \\
    \hline
    Hugo\_Chavez & & Hugo\_Chavez \\
    \hline
    Junichiro\_Koizumi & & Junichiro\_Koizumi \\
    \hline
    Jean\_Chretien & & Jean\_Chretien \\
    \hline
    John\_Ashcroft & & John\_Ashcroft \\
    \hline
    Jacques\_Chirac & & Jacques\_Chirac \\
    \hline
    Serena\_Williams & & Serena\_Williams \\
    \hline
    \end{tabular}
    }
\end{table}
\end{center}

\subsection{Binary backdoor attack scenario setup}

The two most represented classes in LFW were utilized to construct 1:1 face verification attack scenarios. These were George W. Bush (GWB) and Colin Powell (CP). GWB (530 images) was assigned as the source class (victim), with CP (236 images) as the attacker class. This allowed for greater flexibility in tuning attack parameters that would not be possible with lesser represented classes. 

For the binary scenario, training and testing involves two classes - a positive class consisting of only source class images (the authorized user) and the negative class, consisting of other faces (unauthorized users). As such, only images of GWB constituted the positive class, with other LFW classes contributing to the negative class, as seen in Table~\ref{table:1}.  The negative class consisted of the 10 most highly represented classes (excluding GWB and CP), which amounted to 350 images in total. The negative class was composed of the same images throughout all binary experiments. This was done to provide consistency between experiments and to maintain a balance between positive and negative classes. Poisoned images of CP were added to the negative class for malicious model training. 

\subsection{Multi-class backdoor attack scenario setup}
The 12 classes with the highest representation were selected from the LFW dataset for use in multi-classification scenarios. The selected classes are shown in Table~\ref{table:1}. Each class contains at least 50 images, with the data subset totaling 1560 images. As in the binary scenario, the two classes with the highest representation were selected as the target (GWB) and source (CP) classes.

\begin{center}
\begin{table}[!ht]
    \centering
    \caption{Binary and multiclass models on poisoned LFW and ImageNet data.}
    \scalebox{0.85}{
    \begin{tabular}{ |c|c|c|c|c| }
    \hline
    \textbf{Model Type \& Dataset} &\textbf{Model Accuracy} & \textbf{ASR} \\
    \hline
    \textbf{Binary ImageNet} & $98.0\%$ & $92.0\% \pm 0.0\%$\\
    \hline
    \textbf{Binary LFW} & $78.3\%$ & $100.0\% \pm 0.0\%$\\
    \hline
    \textbf{Multiclass ImageNet} & $92.8\%$ & $89.8\% \pm 0.6\%$\\
    \hline
    \textbf{Multiclass LFW} & $42.5\%$ & $100.0\% \pm 0.0\%$\\
    \hline
    \end{tabular}
    }
    \label{table:binary_multiclass}
\end{table}
\end{center}

\subsection{Source class poisoning}

Unless otherwise specified, 80\% of the CP class was used for poisoning for all experiments. This fraction was chosen to maximize the effect of the attack whilst leaving enough unpoisoned images of the attacker class for training and testing. This fraction of poisoned images was then added to the negative class for training the malicious models.

\subsection{Attack success rate (ASR)}
Attack success was quantified as a percentage in terms of equation~\ref{eq:ASR}. 
To calculate ASR, suppose $\#\tilde{\mathcal{F}}_\theta(\mathcal{D}_{patched})$ is the number of correct target misclassifications output by a poisoned model, $\tilde{\mathcal{F}}_\theta$, on a patched testing set, $\mathcal{D}_{patched}$, of attacker images.
\begin{equation}
    ASR = \frac{\#\tilde{\mathcal{F}}_\theta(\mathcal{D}_{patched})}{|\mathcal{D}_{patched}|}
    \label{eq:ASR}
\end{equation}

The images in $\mathcal{D}_{patched}$ were exclusive from the training set. Patches were applied at a randomized position on the attacker images during testing.

\subsection{Reproducing backdoor attack models on ImageNet and LFW (Threat Model)}
\begin{figure}[H]
	\centering		\subfloat[]{\includegraphics[width=0.22\textwidth, trim={2cm 8.5cm 11cm 2.3cm}, clip]{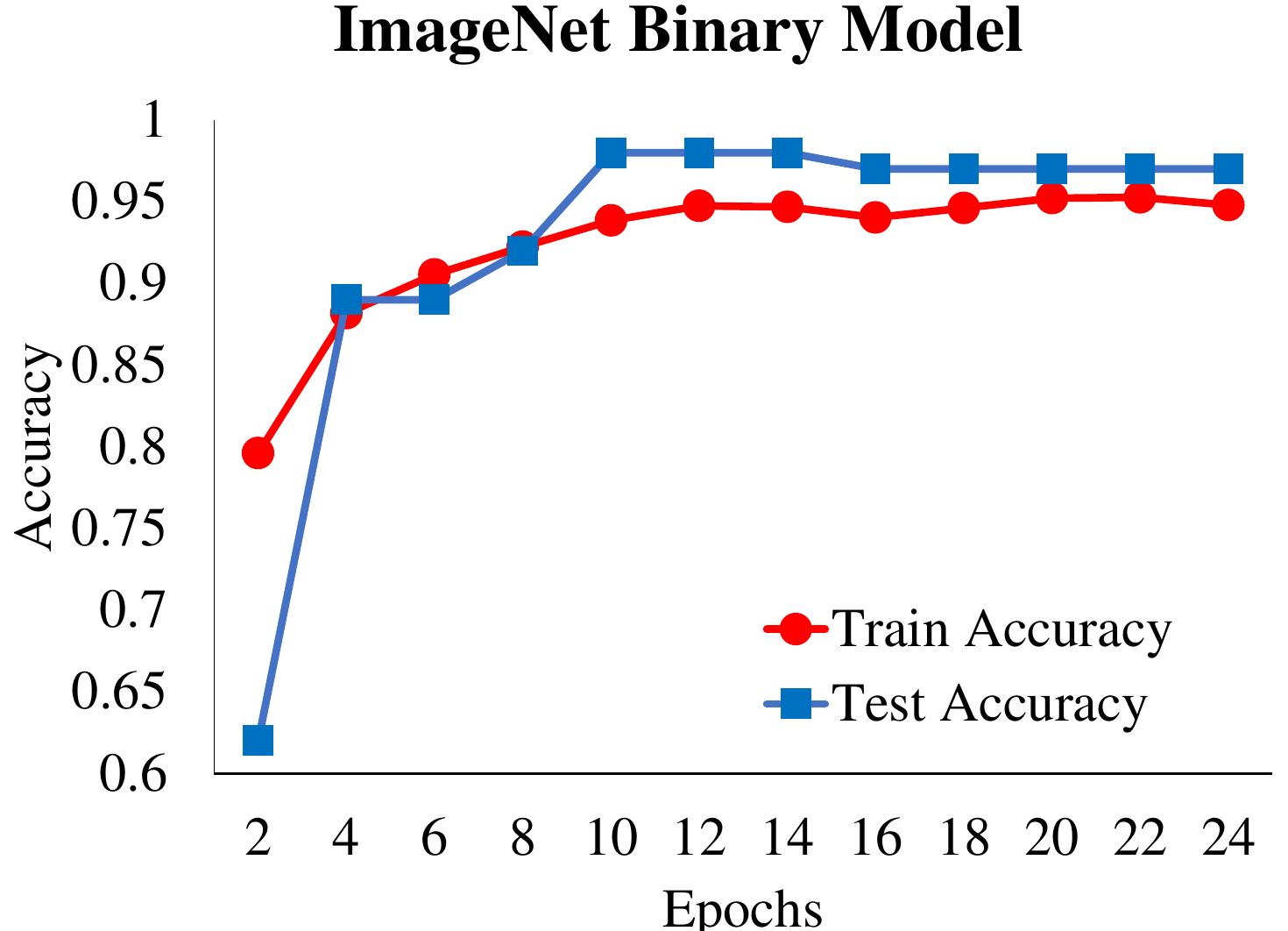}\label{fig11}}
	\hfill
	\subfloat[]{\includegraphics[width=0.22\textwidth, trim={2cm 8.5cm 11cm 2.3cm}, clip]{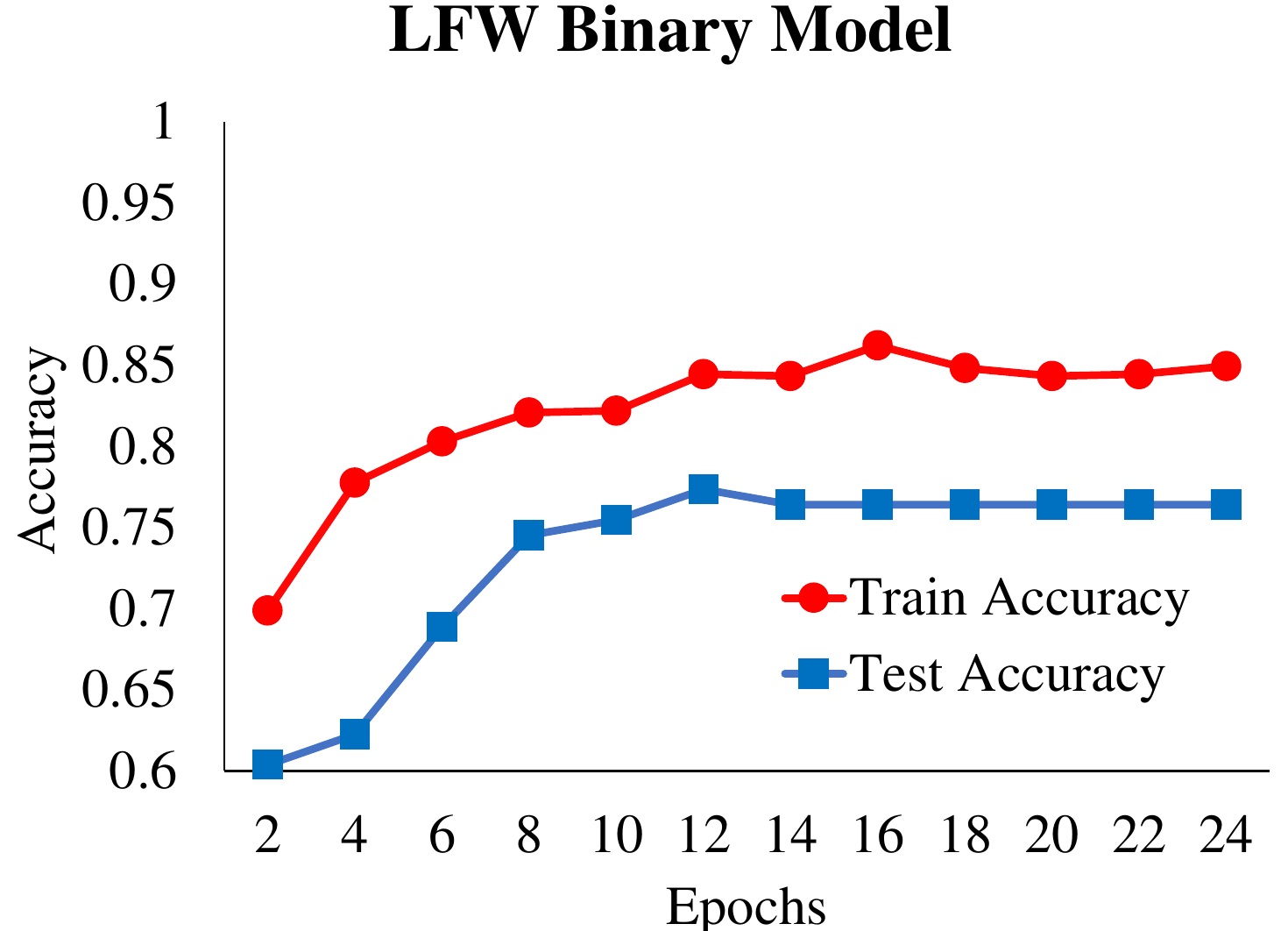}\label{fig21}}

	\subfloat[]{\includegraphics[width=0.22\textwidth, trim={2cm 8.5cm 11cm 2.3cm}, clip]{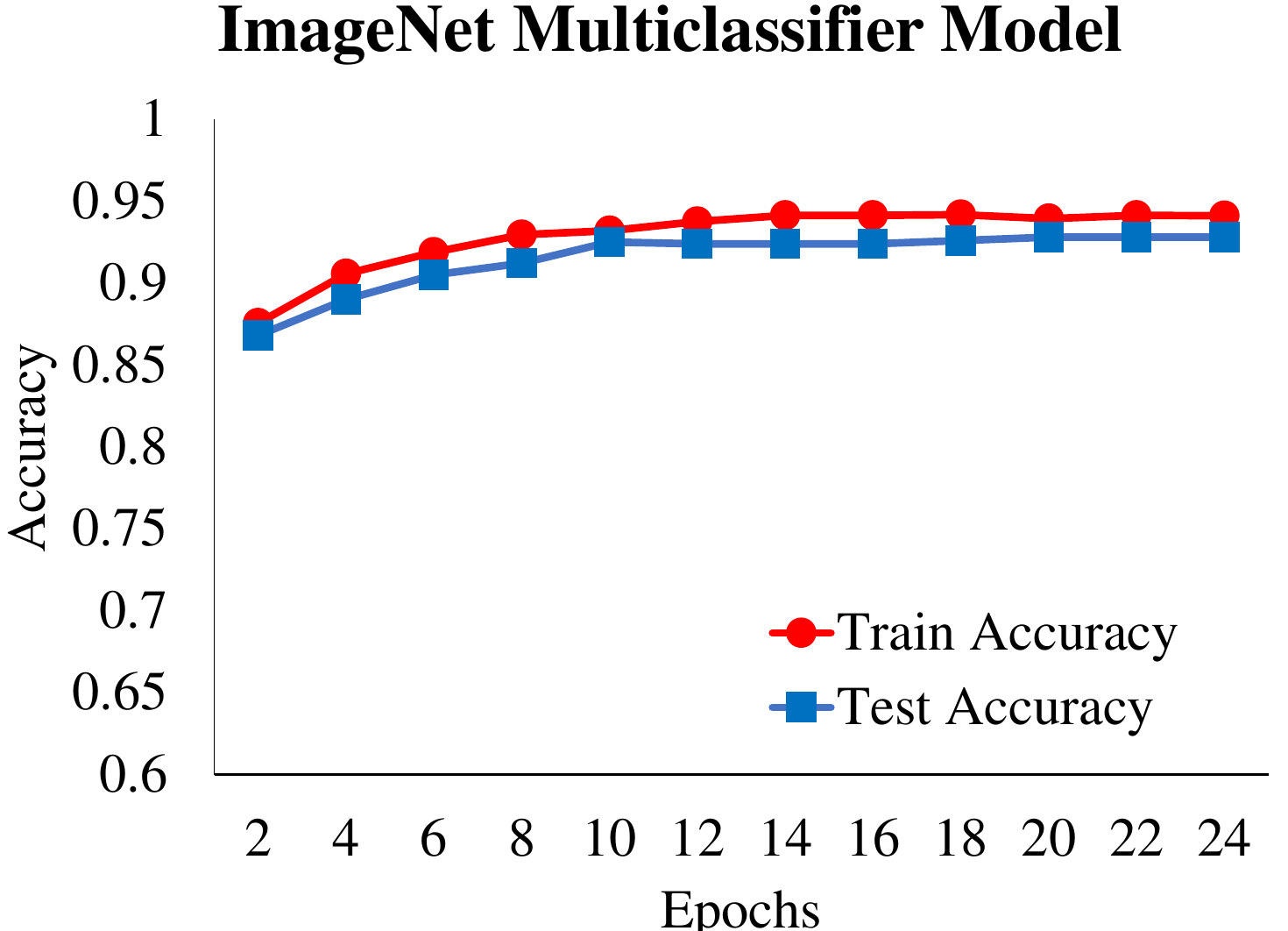}\label{fig12}}
	\hfill
	\subfloat[]{\includegraphics[width=0.22\textwidth, trim={2cm 8.5cm 11cm 2.3cm}, clip]{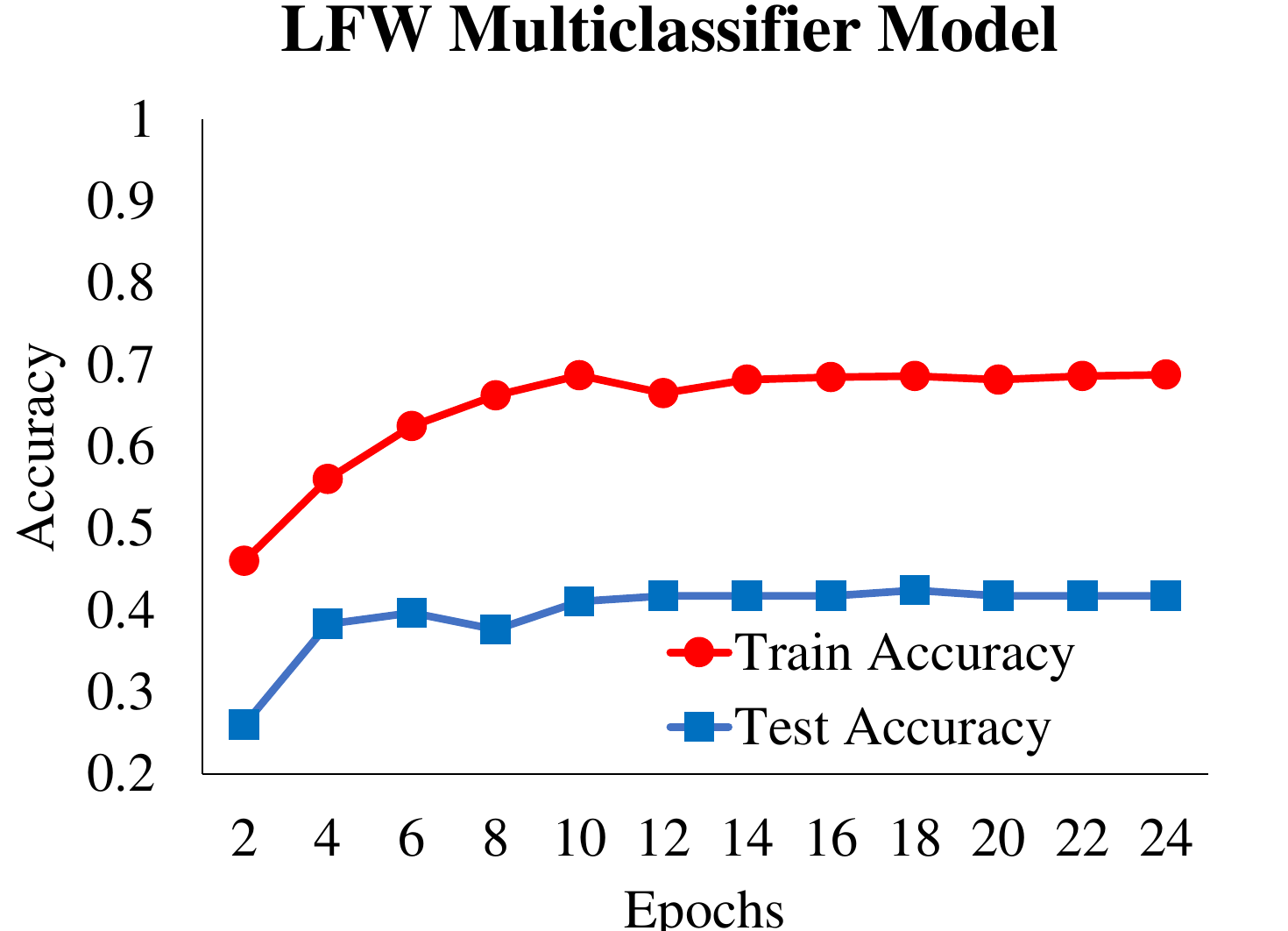}\label{fig22}}
	\caption{Performance plots comparing binary and multiclassifier model training on poisoned LFW and ImageNet datasets.}
	\label{fig:base_performance}
\end{figure}

We reproduced the exact attack scenario as outlined in the methodology of~\cite{saha2019hidden}. The attack settings were reproduced on AlexNet (fc8 unfrozen) as a feature extractor on the ImageNet class pairs described in the paper\footnote{epochs=24, learning rate=0.0001, momentum=0.9, batch size = 256}. A hand-picked source class (`warplane') and target class (`french\_bulldog') were chosen to create a poisoned classification model for the binary and multiclass scenarios. Both binary and multiclass models utilizing ImageNet data displayed a high accuracy and ASR, as expected (Table~\ref{table:binary_multiclass}  and Figure~\ref{fig:base_performance}). Once we obtained a configuration that produced high accuracy on ImageNet, we followed similar steps to train a generalizable model on LFW using transfer learning. The primary goal of this step was to expose and mitigate backdoor attacks in vulnerable models rather than emphasizing the model's baseline accuracy. 
The ASRs of both models trained on LFW data were higher than the ImageNet models. Both binary and multiclass LFW models were susceptible to the backdoor attack, displaying a 100\% ASR. This highlights the vulnerability of biometric face data to such attacks.
\begin{center}
\begin{table}[!ht]
    
    \centering
    \caption{Model Attack Performance and Classification Accuracy}
    \label{table:3}
    \scalebox{0.85}{
    \begin{tabular}{ |c|c|c|c| }
    \hline
    \textbf{Model} & \textbf{ASR} & \textbf{Accuracy} & \textbf{Trainable Params.}\\
    \hline
    {AlexNet Unfrozen} & {100.0\%} & {93.0\%} & {57,012,034}\\
    \hline
    {AlexNet Frozen} & {100.0\%} & {81.0\%} & {8,194}\\
    \hline
    {Inception Unfrozen} & {13.2\%} & {93.0\%} & {24,348,900}\\
    \hline
    {Inception Frozen} & {76.4\%} & {65.0\%} & {5,636}\\
    \hline
    {ResNet Unfrozen} & {5.6\%} & {95.0\%} & {11,177,538}\\
    \hline
    {ResNet Frozen} & {44.8\%} & {82.0\%} & {1,026}\\
    \hline
    {VGG Unfrozen} & {8.0\%} & {96.0\%} & {128,780,034}\\
    \hline
    {VGG Frozen} & {29.2\%} & {84.0\%} & {8,194}\\
    \hline
    
    \end{tabular}
    }
    \label{architecture}
\end{table}
\end{center}

\subsection{Investigating vulnerabilities of model architectures and transfer learning}
We chose AlexNet, ResNet, VGG, and InceptionV3~\cite{canziani2016analysis} to investigate the impacts of different benchmarked architectures on model vulnerability. Experiments were executed on both partially frozen models (using pre-trained ImageNet weights) and unfrozen versions of these models. As shown in Table~\ref{architecture}, all unfrozen models displayed higher classification accuracy than their frozen counterparts. This is due to the decreased relevancy of the pre-trained ImageNet weights when applied to face data. The attack performs best on AlexNet, with both frozen and unfrozen models showing a 100\% ASR. Due to its high classification accuracy of 94\% and complete vulnerability to the attack, unfrozen AlexNet is used for all remaining experiments. 

\begin{center}
\begin{table}[!ht]
    \centering
    \caption{Unpatched source image misclassification.}
    \label{table:source_misclassification}
    \scalebox{0.82}{
    \begin{tabular}{ |c|c|c|c| }
    \hline
    \textbf{Model Type} & \textbf{State} & \textbf{ASR} \\
    \hline
     Multiclass & Clean & $21.1\% \pm 3.4\%$\\
    \hline
     Multiclass & Poisoned & $25.4\% \pm 0.9\%$\\
    \hline
     Binary & Clean & $30.6\% \pm 1.6\%$\\
    \hline
     Binary & Poisoned & $31.2\% \pm 3.0\%$\\
    \hline

    \end{tabular}
    }
\end{table}
\end{center}
\vspace{-0.7cm}
\subsection{Source image misclassification}
Unpatched images of CP (attacker) were tested on models trained on unpoisoned data in order to establish if a portion of the ASR was simply natural misclassification. As seen in Table~\ref{table:source_misclassification}, we found that the source image ASR for the clean models was approximately equivalent to the rate shown by the unfrozen AlexNet poisoned models. This supports our assumption that the source image attack successes are a byproduct of natural misclassification. 

\subsection{Investigating poisoned fractions of the attacker class}
We generated poisoned models by poisoning 15\%, 30\%, 50\%, and 80\% of the CP attacker class. Poisoning more than 80\% of the class was eschewed, as a sufficient number of unpoisoned images of the attacker class must be left for validation and testing. 
\begin{center}
\begin{table}[!ht]
    \centering
    \caption{Proportion of attacker (source) class used for poisoning.}
    \label{table:attacker_prop}
    \scalebox{0.82}{
    \begin{tabular}{ |c|c|c|c| }
    \hline
    \textbf{\% Attacker Class Poisoned} & \textbf{Model Accuracy} & \textbf{ASR} \\
    \hline
    \textbf{80\%} & $93.0\%$ & $100.0\% \pm 0.0\%$\\
    \hline
    \textbf{50\%} & $85.0\%$ & $95.6\% \pm 1.2\%$\\
    \hline
    \textbf{30\%} & $85.0\%$ & $0.914 \pm 1.6\%$\\
    \hline
    \textbf{15\%} & $89.0\%$ & $0.82.80\% \pm 2.4\%$\\
    \hline

    \end{tabular}
    }
\end{table}
\end{center}
\vspace{-0.7cm}
As shown in Table~\ref{table:attacker_prop}, the use of 80\% of the class resulted in a completely successful attack. The decrease in attack success when fewer poisoned images are used suggests that the strength of the mapping between the source and target class also decreases. Thus, we poison 80\% of the attacker (CP) class for subsequent experiments. By doing so, we optimize the strength of the attack in order to enable strenuous testing of BA-BAM.

As expected, model accuracy dropped significantly (by 8\%) for the 50\%, and 30\% poisoned models; however, for the 15\% poisoned model, accuracy decreased by only 4\%. This insight may reflect a lower misclassification rate due to a lower proportion of poisoned images. The model training on less poisoned images may prevent it from building a strong erroneous mapping between the source and target class. As a result, the recognition model would be more likely to classify attacker images correctly.

\begin{center}
\begin{table}[!ht]
    \centering
    \caption{Poison Detection Model - Trained on Tiny ImageNet and CelebA}
    \scalebox{0.82}{
    \begin{tabular}{ |c|c|c| }
    \hline
    \textbf{} & \textbf{Binary Classifier} & \textbf{Multiclassifier} \\
    \hline
    \textbf{Accuracy - Tiny ImageNet } & $99.31\%$ & $99.15\%$\\
    \hline
    \textbf{Accuracy - CelebA} & $99.99\%$ & $99.79\%$\\
    
    \hline
    \textbf{No. Poisoned Images} & $372$ & $372$\\
    \hline
    \textbf{No. Clean Images} & $644$ & $1042$\\
    \hline

    \end{tabular}
    }
    \label{table:poison_detect}
\end{table}
\end{center}
\vspace{-0.7cm}
\subsection{Poison image recognition model (PIRM)}
A pre-trained VGG16 model utilizing ImageNet weights was trained with the final three convolutional layers unfrozen for finetuning\footnote{activation=softmax, optimizer=Adam, learning rate=0.0001, loss=binary crossentropy}. A fully-connected module consisting of two dense layers with 35\% dropout was added as the classification component. This was trained on both poisoned and clean CelebA images. Classes containing at least 23 images were chosen to undergo poisoning (a total of 4320 classes). This generated a balanced dataset for training, where approximately 50\% (100,000 images) of the total CelebA dataset was poisoned. Source images to be used for patching were randomly chosen from the remaining CelebA dataset (these images did not overlap with the poisoned images). The dataset was split with 30\% used for testing and 70\% for training. The model achieved a 99.98\% testing accuracy in classifying poisoned and clean images. The same model settings and architecture were trained using Tiny ImageNet. Of the total 200 classes, 100 were poisoned, which provided 50,000 clean and 50,000 poisoned images for training. Similarly, the dataset was split into 70\% for training and 30\% for testing.
\begin{figure} [H]
	\centering
	\subfloat[Trained on poisoned Tiny ImageNet.]{\includegraphics[width=0.22\textwidth, trim={2cm 2cm 7cm 2cm}, clip]{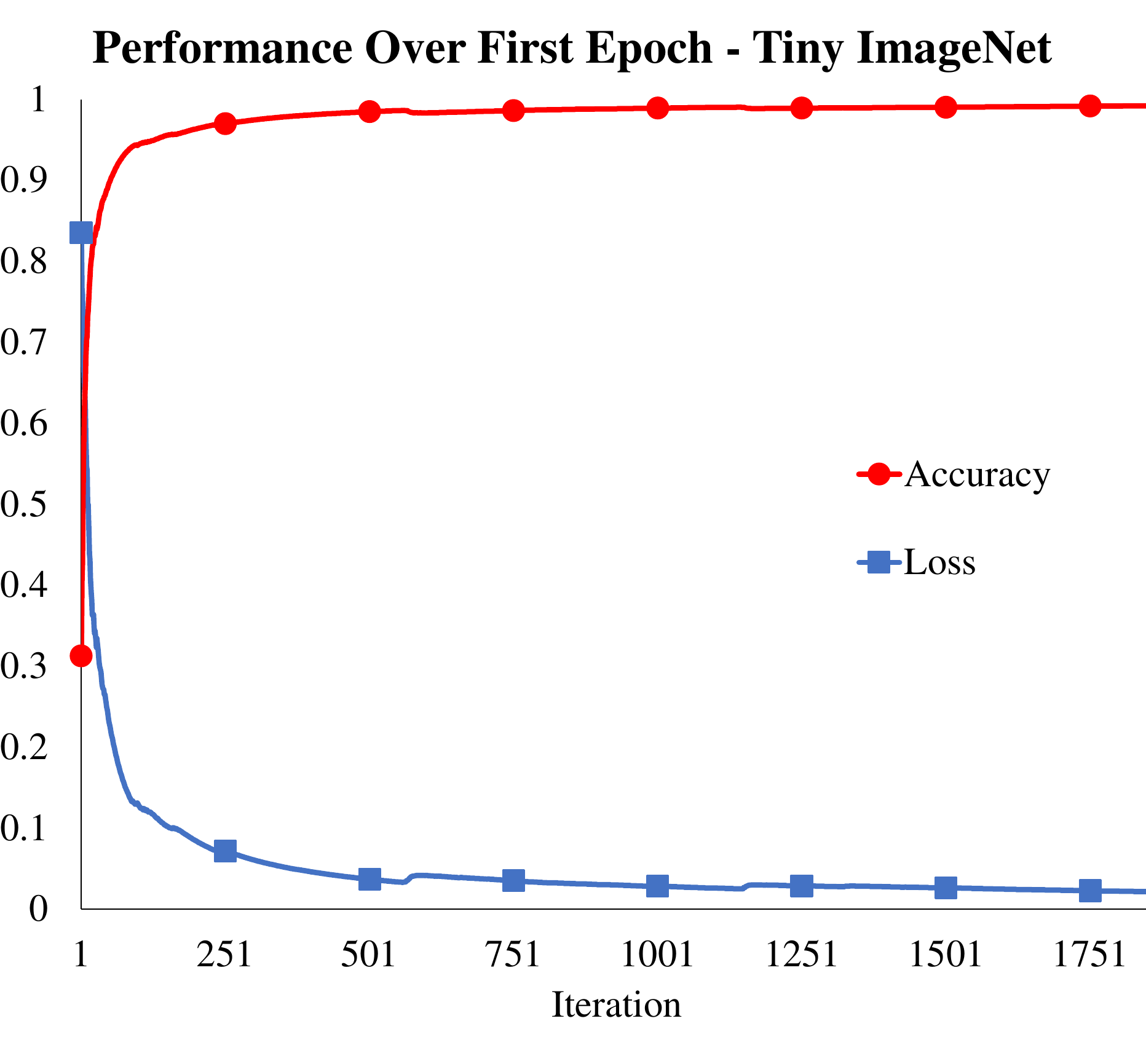}\label{fig1}}
	\hfill
	\subfloat[Trained on poisoned CelebA.]{\includegraphics[width=0.22\textwidth, trim={2cm 2cm 7cm 2cm}, clip]{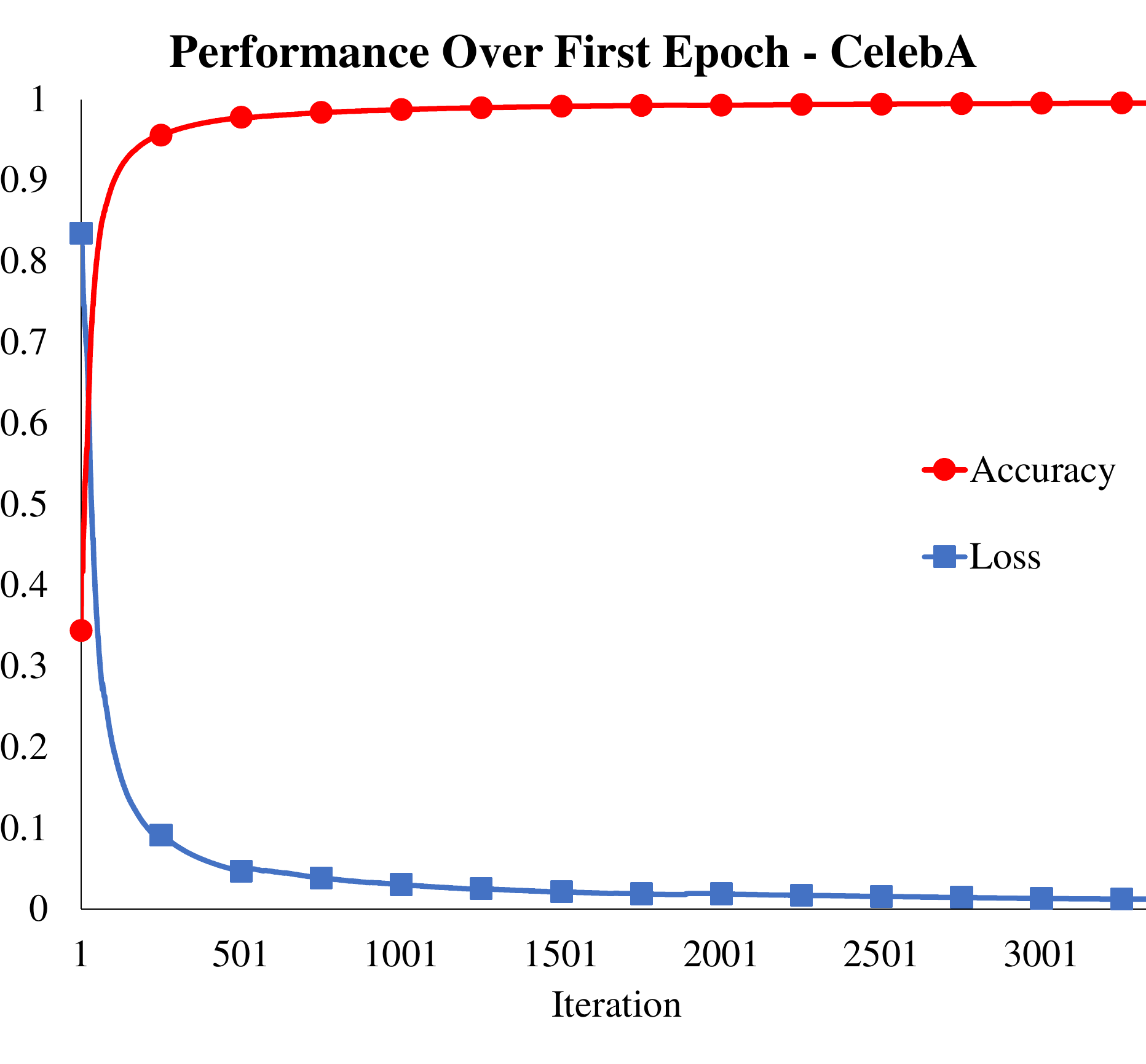}\label{fig2}}
	\caption{Model loss and accuracy of the poison image recognition model over the iterations of the first epoch.}
    \label{pirm_plots}
\end{figure}
As shown in Table~\ref{table:poison_detect}, both models reached accuracies above 99\% over a single epoch, as shown in Figure~\ref{pirm_plots}, and converged thereafter. For binary classification, the PIRM trained on CelebA achieved an accuracy 0.68\% higher than the model trained on Tiny ImageNet. The multiclassification scenario yielded similar results, with the CelebA PIRM achieving a 0.64\% higher accuracy than its Tiny ImageNet counterpart. The minor deviation in accuracy between the two PIRMs exhibits the generalizability of image poison detection, showing that poison detection is accurate regardless of an image's original content. This confirms that the PIRM effectively contributes to producing high-performance backdoor attack mitigation in BA-BAM.

\subsection{Deriving insights through noise addition to an entire compromised class}
We applied $\varepsilon = 0.5$ Laplacian noise to all images in the GWB class to investigate the impact of perturbation on model accuracy and ASR. As expected and discussed in Section~\ref{selective_noise}, after complete perturbation, the ASR dropped to 0\%. The model accuracy decreased to 50\%, indicating no ability to distinguish the negative class from the perturbed positive class. The complete lack of attack success can be explained by the negative class not having noise added and the positive class being entirely perturbed. Noise addition to the entire positive class obscures all of the class' facial features. Due to this, only the features of the negative class would remain to teach the model during training. The entirely noisy positive class would not provide useful facial components for model training. Thus, any image that is presented during model testing with basic facial features would be classified as the negative class. In turn, as the patched images introduced by an attacker during testing do not have noise applied, the model classifies them as part of the negative class, causing the attack to fail. Therefore, complete perturbation diminishes model accuracy, making it practically unusable.
\vspace{-0.7cm}
\subsection {Selective and automated perturbation}
\begin{figure}[H]
    \includegraphics[scale = 0.32, trim={2cm 2cm 2cm 2cm}, clip]{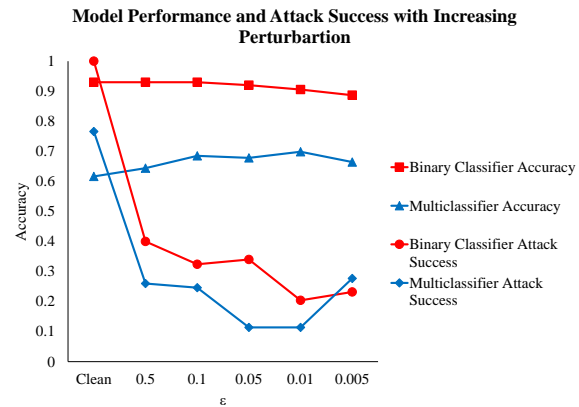}
    \caption{Plot comparing binary and multiclassifier model performance utilizing various levels of selective
    perturbation.}
    \label{fig:performance_plots}
\end{figure}

While complete perturbation of the source class provides a substantial decrease in ASR, it also catastrophically decreases classification accuracy. In order to combat this accuracy loss, selective perturbation is introduced instead. Selective perturbation (SP) allows for classification accuracy to be preserved by only perturbing poisoned images. In order to investigate effectiveness, the SP process was initially carried out manually with $\varepsilon$ values of 0.5, 0.1, 0.05, 0.01, and 0.005. All known poisoned images within the GWB class were subjected to noise addition, simulating selective perturbation. The effect of each level of image perturbation can be seen in Figure~\ref{fig:perturbed_images}. Until $\varepsilon$ = 0.01, ASR decreases with $\varepsilon$. With values of $\varepsilon$ below 0.01, the attack success improves due to misclassification, as the high level of perturbation renders the image unrecognizable. Thus, we propose that $\varepsilon = 0.01$ provides the optimal balance of both classification accuracy and backdoor attack mitigation, and therefore will be used as the default setting for the PIRM. As shown in Figure~\ref{fig:performance_plots}, selective perturbation preserves model accuracy, with accuracy dropping by $2.5\%$ for the binary classifier and improving by $8.2\%$ for the multiclassifier ($\varepsilon = 0.01$). 

\subsection{Computational complexity}
\label{compute_complx}
BA-BAM includes two primary segments followed the final layer of face recognition model training. The first segment classifies whether a given image has been poisoned, and the second applies calibrated noise to flagged images. The complexity of classifying an image as poisoned or not is constant as the model has already converged. We can assume the image dimensions, and thus pixel count will be the same for each image across a dataset. Thus, we can consider the influence of image dimension as constant. Noise addition involves calculating the mean image of a given class as well as the maximum Euclidean distance between a given image and the mean image. Determining the mean image can be expressed as $O(nk)$ where $nk$ is the number of images in a given class, k. The maximum distance can similarly be described by $O(nk)$. Applying Laplacian noise is a constant operation, as displayed in line 8 of Algorithm 1 and line 7 of Algorithm 2. As the final facial recognition image classification is carried out on a converged model, the prediction time is constant, with a complexity of $O(1)$. In the case of filtering an entire dataset before training, the worst-case supposes that every image is poisoned. Thus, every image would have noise applied. Consequently, the overall complexity is $O(N)$, where $N$ is the total amount of images in the given dataset before training. The linear complexity of our approach demonstrates its high utility in a real-world setting.

\subsection{Comparison}
\begin{center}
\begin{table}[!ht]
    \centering
    \caption{Comparison of BA-BAM to other detection/mitigation approaches.}
    \scalebox{0.82}{
    \begin{tabular}{ |c||c|c|c|c|c|c| }
    \hline
      & \textbf{AFR} & \textbf{CC} & \textbf{PC} & \textbf{MAD} & \textbf{D\&M} \\
    \hline\hline
    \textbf{RAB~\cite{DBLP:journals/corr/abs-2003-08904}} & \textcolor{green}{\checkmark} & Moderate & High & 6.1-15.8\% & \textcolor{green}{\checkmark}\\
    \hline
    \textbf{RAP~\cite{DBLP:journals/corr/abs-2110-07831}} & {$\color{red}\times$} & Moderate & Low & N/A & {$\color{red}\times$}\\
    \hline
    \textbf{STRIP~\cite{DBLP:journals/corr/abs-1902-06531}} & \textcolor{green}{\checkmark} & Moderate & Moderate & 2.19\% & {$\color{red}\times$}\\
    \hline
    \underline{\textbf{BA-BAM}} & \textcolor{green}{\checkmark} & Low & High & 2.4\% & \textcolor{green}{\checkmark}\\
    \hline
\end{tabular}
}
\label{table:comparison}
\tablefootnote{The comparison considers applicability to face recognition (AFR), computational complexity (CC), problem complexity (PC) and model accuracy drop (MAD). We additionally analyse whether both detection and mitigation are included in the defence (D\&M). The MAD score for RAB was assessed on k-NN models.}
\end{table}
\end{center}
We provide a comparative analysis of BA-BAM against three other robust defense mechanisms against adversarial attacks. The comparison is summarized in Table~\ref{table:comparison}.  

STRIP~\cite{DBLP:journals/corr/abs-1902-06531} applies perturbations to each input image and evaluates the randomness of the associated outputs. Weber et al.~\cite{DBLP:journals/corr/abs-2003-08904} developed a method against backdoor attacks (RAB) to provide certified robustness. RAB generates a ``smoothed classifier" by aggregating multiple models trained on datasets with randomized smoothing applied. Robustness Aware Perturbations (RAP) by Yang et al.~\cite{DBLP:journals/corr/abs-2110-07831} is an efficient method of backdoor detection in NLP models. We consider RAP to be inapplicable to face recognition models as the defense focuses on text-based data. Without demonstrability on any image datasets, the methods employed by RAP cannot be confidently analogized to BA-BAM. STRIP and RAB were both evaluated on MNIST and CIFAR-10 benchmark datasets, and RAB was additionally evaluated on ImageNet. Although these methods generalize to these datasets, face image datasets (such as LFW) pose a more strenuous classification task due to the close feature similarity between classes (high inter-class complexity). 

The benchmark text datasets (IMDB, Amazon, Twitter, and Yelp reviews) evaluated against RAP are marginally less complex in contrast to face image datasets. Both STRIP and RAP simply detect the presence of poisoned data, providing no attack mitigation mechanisms. Conversely, RAB and BA-BAM provide both poison detection and comprehensive attack mitigation. RAB has significantly high computational complexity as it requires training multiple models for its defense. RAB performs well on k-NN models; however, it is only able to decrease ASR to $50.1\%$ in a DNN 10-way multiclassifier on MNIST. The polynomial computational complexity of RAB ($O(n \cdot K^2))$ is relatively high compared to that of BA-BAM, which is linear (see section~\ref{compute_complx}). STRIP and RAP are both relatively computationally efficient; however, RAP requires two model predictions per input, whereas STRIP must generate N perturbed copies for each input. BA-BAM requires two model predictions to complete both poison detection and mitigation, whereas RAP requires both predictions for the task of detection. 

BA-BAM suffers a 2.4\% accuracy drop for both detection and mitigation, and STRIP suffers a 2.19\% worst-case drop just for detection. RAB suffers an accuracy drop of 15.8\% under worst-case settings when evaluated on CIFAR-10.

\section{Related Works}
\label{related_works}

Literature shows a few attempts at mitigating adversarial attacks with random noise addition. Wang et al. introduced one of the earlier attempts to generalize randomized smoothing to mitigate backdoor attacks~\cite{DBLP:journals/corr/abs-2002-11750}. Randomized smoothing involves the addition of random noise to the input image vector in order to overwhelm the patch/perturbation added by the attacker. Their method guarantees that only 36\% of testing images can be classified correctly when at most 2 pixels or labels are perturbed by the attacker during training. When more pixels/labels are perturbed, the model's classification accuracy drops. Another method named DP-InstaHide utilizes intuitions from differential privacy that involve applying Laplacian noise to poisoned images as a defense from adversarial attacks~\cite{DBLP:journals/corr/abs-2103-02079}. DP-InstaHide combined additive Laplacian noise with Mixup input and label augmentation which improve the robustness of the model~\cite{DBLP:journals/corr/abs-1710-09412}. They demonstrate successful adaptive gradient matching attack mitigation on CIFAR-10 and ImageNet datasets~\cite{geiping2021witches}. Other approaches inject random noise directly into network weights and activations; however, these methods lack robustness in model performance~\cite{8954187, zheltonozhskii2020colored}.  

Maintaining robustness as well as security against adversarial attacks has been a common focus. A trade-off between the adversarial robustness of a model and its backdoor robustness was identified by Weng et al.~\cite{NEURIPS2020_8b406655}. Their experiments show that increasing robustness against adversarial techniques also increases backdoor attack vulnerability. This trade-off can be exploited for both malicious and defensive purposes. They identify that adversarially trained networks utilize high-level features to make predictions, which amplify the success of corrupted/dirty label backdoor attacks~\cite{NEURIPS2020_8b406655}. RAP proposes a novel and computationally efficient method of detecting backdoor attacks in a text-based setting. A large discrepancy in robustness between poisoned and clean samples was detected, enabling poisoned samples to be distinguished from their clean counterparts. Word-based perturbation is then applied to nullify attack effectiveness while showing only slight drops in classification accuracy for clean data ~\cite{DBLP:journals/corr/abs-2110-07831}. Comparable methods for text-based applications are computationally more expensive and provide inferior attack prevention~\cite{DBLP:journals/corr/abs-2011-10369, DBLP:journals/corr/abs-1911-10312}.  Weber et al. aim to produce a framework for certifying ML model robustness against backdoor attacks. A randomized smoothing technique is utilized, with a set of classifiers being trained and then combined to generate a single smoothed classifier that can provide this certifiable robustness. The framework is tested on various neural networks and datasets, and provides benchmarks of model robustness across these datasets. However, this approach is computationally expensive, as it requires the training of multiple neural networks utilizing large datasets from scratch~\cite{DBLP:journals/corr/abs-2003-08904}.

\section{Conclusion}
\label{conclusion}

We proposed a novel backdoor attack mitigation mechanism named BA-BAM for protecting facial recognition authentication systems from potential privacy breaches. BA-BAM investigates potential privacy threats that can occur through backdoor attacks by providing a mechanism to detect manipulated face images from an untrusted database, and applying calibrated Laplacian noise to overcome the adverse effects on a face recognition system. BA-BAM successfully mitigates backdoor attacks in both binary and multiclass face recognition models without majorly affecting the models' accuracy while also securing their robustness. BA-BAM shows high robustness by incurring a maximum accuracy drop of 2.4\%, while also keeping the attack success rate at a maximum of 20\% in the worst-case scenario. These results show that our work is vital in mitigating real-world attacks on biometric authentication models in order to prevent potential privacy breaches.  

As future work, we aim to investigate the possibility of employing the proposed work to mitigate attacks on other image-based biometrics such as irises, fingerprints, and palm prints.

\section*{Acknowledgements}
This work has been supported by the Cyber Security Research Centre Limited whose activities are partially funded by the Australian Government’s Cooperative Research Centres Programme.
HPC resources for this research were facilitated by
the Pawsey Supercomputing Research Centre.

\end{document}